\documentclass[letterpaper, 10 pt, conference]{ieeeconf}  

\IEEEoverridecommandlockouts                              

\usepackage{epsfig}
\usepackage{makecell}
\usepackage{subcaption}
\usepackage{xcolor}
\usepackage{ragged2e}
\usepackage{float} 

\usepackage{graphicx}
\usepackage{amsmath}
\usepackage{adjustbox}
\usepackage{amssymb}
\usepackage{booktabs}
\usepackage{tabularx}
\usepackage{array}
\usepackage{makecell}

\usepackage{bm}

\title{\LARGE \bf
Semantic Trajectory Data Mining with LLM-Informed POI Classification}

\author{Yifan Liu, Chenchen Kuai, Xishun Liao*, Haoxuan Ma, Brian Yueshuai He, and Jiaqi Ma
\thanks{The authors are with the UCLA Mobility Lab under the Department of Civil and Environmental Engineering, University of California, Los Angeles, Los Angeles, USA.}
\thanks{*Corresponding author: xishunliao@ucla.edu}
}

\begin{document}

\maketitle
\thispagestyle{empty}
\pagestyle{empty}

\begin{abstract}
Human travel trajectory mining is crucial for transportation systems, enhancing route optimization, traffic management, and the study of human travel patterns. While previous studies have primarily focused on spatial-temporal information, the integration of semantic data has been limited, leading to constraints in efficiency and accuracy. Semantic information, such as activity types inferred from Points of Interest (POI) data, can significantly enhance the quality of trajectory mining. However, integrating these insights is challenging, as many POIs have incomplete feature information, and learning-based POI algorithms require the integrity of datasets to do the classification. In this paper, we introduce a novel pipeline for human travel trajectory mining, annotating GPS trajectories with POIs and visit purpose. Our approach first leverages the strong inferential and comprehension capabilities of large language models (LLMs) to link POI with activity types and then uses a Bayesian-based algorithm to infer activity for each stay point in a trajectory. In our evaluation using the OpenStreetMap POI dataset, our approach achieves a 93.4\% accuracy and a 96.1\% F-1 score in POI classification, and a 91.7\% accuracy with a 92.3\% F-1 score in activity inference.


\end{abstract}

\maketitle
\section{Introduction}



The rapid development of Internet-of-Thing (IoT) technology and applications facilitates the evolution of intelligent transportation systems and brings a new era of data collection~\cite{10480881, 10356725}. This flourish yields an abundance of trajectory data from a wide range of connected devices and allows us to explore human travel behavior with additional detail and accuracy. The advent of GPS-enabled mobile devices has revolutionized the tracking of individuals, vehicles, trains, and even animals through the collection of digital traces or trajectories~\cite{10131965, 10354062, pasquaretta2021analysis, 10554659}, enabling the study of travel behaviors. 

However, current GPS-based datasets contain only spatial-temporal information, which limits their ability to fully address the complexities of mobility behavior studies, especially when it comes to understanding human mobility. For instance, these datasets do not capture the travel intentions behind each stay point, as human travel often involves specific purposes and underlying interdependencies. Therefore, to model human mobility patterns more accurately, there is a growing demand to integrate GPS-based trajectory data with semantic information. As pointed out in ~\cite{shen2014review}, this integration facilitates applications in traffic management, disease analysis, and human movement studies.



An integrated dataset combining location trajectories with semantic information would ideally include not only the spatial-temporal data but also information about nearby Points of Interest (POIs), the purpose of visits, and other contextual details. Such a dataset would provide a more comprehensive view of human mobility patterns, allowing for deeper insights into travel behaviors and motivations.

To create such integrated datasets and effectively mine trajectory data, accurate POI classification is crucial. The effectiveness of trajectory mining heavily relies on the accuracy of POI classification, which is also crucial for applications like navigation, local searches, and analyzing human travel patterns~\cite{teusch2023systematic, ruta2012semantic, thonhofer2023infrastructure}. A reliable POI classification algorithm can notably enhance the usability, functionality, and effectiveness of these applications by ensuring the quality of POI data categorization. Despite their importance, existing learning-based classification methods suffer from considerable variability within open-source datasets like OpenStreetMap (OSM)~\cite{haklay2008openstreetmap}, and discrepancies widen when multiple data sources are considered. This inconsistency not only complicates schema alignment but also significantly limits the amount and quality of usable data. Efforts to synchronize these disparate data schema are time-consuming and often result in loss of data granularity and reliability. 

In this paper, we introduce a data mining frame work to annotate trajectory with semantic information. Compared to existing literature, the main contributions of this paper are:

\begin{itemize}
\item We introduce a novel data mining framework to annotate trajectories with activities, which integrates LLM-based POI classification with a probabilistic activity inference algorithm. This semantic annotation of time series data bridges spatial-temporal and natural language analyses in trajectory mining, opening new avenues for mobility research.
\item The proposed framework is adaptable across various regional POI or trajectory datasets without additional training. It effectively handles POI classification even with open-source datasets that have incomplete data, eliminating the need for high data integrity.
\item Leveraging the semantic information inferred from POIs provided by LLM outputs, our activity inference algorithm achieves precise point-level inference. To our knowledge, this is the first application of LLM to POI classification.

\end{itemize}

\section{Literature Review}
\begin{figure*}[t]
    \centering
    \medskip
    \includegraphics[width=1.0\linewidth]{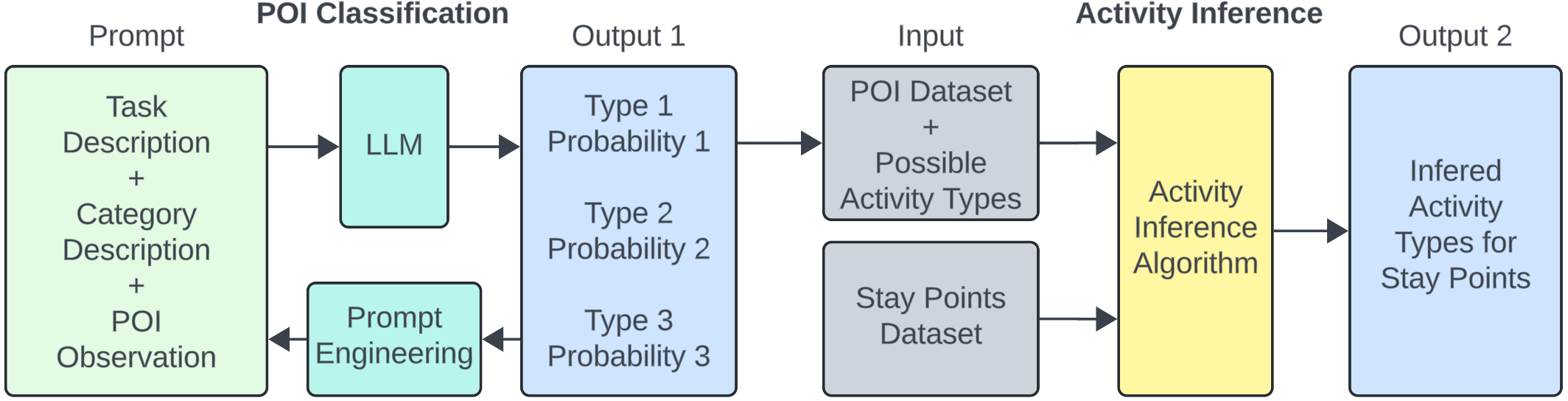}
    \caption{Architecture of the proposed framework from POI classification (left) to activity Inference (right).}
    \label{fig:overall_architecture}
    \vspace{-2mm}
\end{figure*}
\subsection{POI Classification}

The classification of POIs involves assigning categories to each observation based on location types (restaurants, schools) or associated activities (shopping, working). Learning-based methods like SVM initially addressed this by leveraging additional textual information, achieving up to 73\% accuracy on the Yelp dataset with 20 different categories~\cite{choi2014poi}. Integrating spatial, textual, and property features further enhanced classification, with methods like k-Nearest Neighbors achieving up to 91\% top-10 accuracy in the OSM dataset with 14 categories~\cite{zhou2020poi}.

As the number of features expands, assessing feature importance for classification has become crucial. Methods like Linear Discriminant Analysis, kNN, and Random Forest are used to evaluate feature significance in classification tasks~\cite{milias2021assessing}. Research shows that feature importance varies by geographical location and category, complicating the creation of a universally effective algorithm for diverse datasets.

Previous methods predominantly used learning-based algorithms, improving performance via feature extension and data from external sources. In practice, missing features and the challenge of synchronizing diverse data sources complicate classification. Labeling variability, such as OSM datasets from Egypt that feature names exclusively in Arabic, limits the effectiveness of these algorithms.

These challenges highlight the need for more robust and adaptable algorithms. LLMs trained with diverse data including internet content, excel in various applications with their strong generalization and inference capabilities~\cite{zhao2023surveylargelanguagemodels}. Our approach utilizes LLMs to address these issues, providing an efficient method for classifying POIs across diverse datasets without requiring complete data or additional feature extensions.

\subsection{Activity Inference}

Trajectory data from GPS devices documents daily trips as sequences of geo-coordinates, revealing insights into human activities like shopping or dining. However, inferring trip purposes is challenging due to GPS inaccuracies and the vague information in POI datasets \cite{bhattacharya2015matching}.

Studies have used spatial and temporal features from trajectories to infer travel behaviors. While rule-based methods accurately predict mandatory activities like home and work due to strong periodicity, they struggle with complex non-mandatory activities like social visits \cite{10382155, alexander2015rule}. To improve inference, some methods integrate POI data, using GNN and Dynamic Bayesian Networks to better predict activity types, though precision remains an issue due to the coarse data \cite{liu2023Graph, meng2017DBN}.

To address these issues, our model is designed to accurately infer activity types using point-level POIs, rather than broader zonal data. Leveraging prior knowledge of activity temporal distributions improves accuracy significantly. We overcome these challenges with a Bayesian-based algorithm that incorporates the context of nearby POIs.

\section{Methodology}

The architecture of our trajectory mining framework is illustrated in Fig. \ref{fig:overall_architecture}. We first perform POI classification by reformulating the task description and candidate activity types into natural language representations. This information is then combined with the POI observations to determine the three most likely activity types for each POI using LLM. Meanwhile, we engage in prompt engineering to refine the quality of outputs based on the model’s performance. Following this, we integrate stay point~\cite{Jiang2016TimeGeo} information and implement a Bayesian-based inference algorithm to accurately associate each stay point with potential POIs and activity types in the human trajectory data.

\subsection{Problem Formulation}
Given the POIs and stay points from trajectory data, the task is to infer activity types for stay points in the trajectory data, and the mathematical expressions and definitions are provided:

\textbf{\textit{POI}}: The point of interest observations are denoted as $\mathcal{P} = \{p_1, p_2,..., p_I\}$, where each observation $p_i$ consists of $\langle \text{Name}, \text{Lon}, \text{Lat}, f_1, f_2 ... f_n \rangle$, indicating the name, longitude, latitude and available features ($f_n$) of the POI.

\textbf{\textit{Trajectory}}: The trajectory data traces are denoted as $\mathcal{T}=\{Traj_1, Traj_2, ..., Traj_N\}$.

\textbf{\textit{Stay Point}}: For each trajectory of person $n$, the stay points are defined as locations where the person remains stationary for a duration of time. A trajectory could have multiple stay points $Traj = \{S_{1},S_{2},...,S_{K}\}$. Each stay point contains information of $(t_S, t_E, $Lon$, $lat$)$, representing the start time, end time, longitude, and latitude of the stay point.

Given information of each stay points $S$ and POI records $\mathcal{P}$, the goal is to annotate each stay point with an activity $A$ for the whole trajectory dataset $\mathcal{T}$. 

\subsection{POI Classification}
\begin{figure}
    \centering
    \medskip
    \includegraphics[width=0.9\columnwidth]{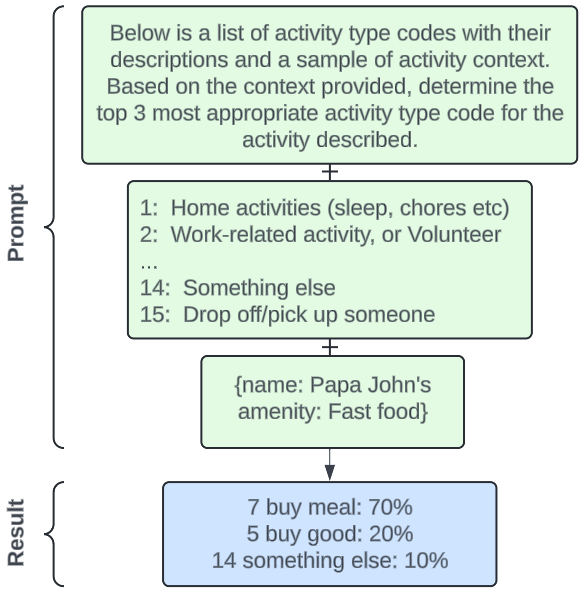}
    \caption{Example of the prompt and output from LLM.}
    \label{fig:prompt_example}
    \vspace{-2mm}
\end{figure}
Different from learning-based algorithms that derive relationships between POI features and categories from labeled data samples, LLM leverages its powerful inferencing capabilities to respond to open-ended text-based questions without requiring additional training. In our approach, we reframe the classification task into a text-based question format where LLMs excel and design efficient prompts that precisely describe the problem, allowing the LLM to interpret and classify POIs effectively based on the context provided in the prompts.

As shown in Fig. \ref{fig:prompt_example} as an example, the prompt is composed of three distinct components:

\textbf{\textit{Task Description}}: Outlines the POI classification task, detailing input data and the desired output structure. This sets the framework for the LLM’s approach to data analysis and classification.

\textbf{\textit{Category Description}}: Provides a detailed description of each target category, including definitions and examples. This helps guide the LLM in accurately classifying POIs into relevant categories.

\textbf{\textit{POI Observation}}: Describes dataset features and their values in natural language format, ensuring the model could fully understand and process each POI’s characteristics as described in the data.

Notably, prompt fine-tuning can be utilized to meet specific needs when working with diverse datasets or various category sets, as LLMs are not limited by text input formats. For instance, in the Egyptian POI collection where some POI names are provided only in Arabic, we can enhance the model's understanding by explicitly stating in the prompt like ``some names in the `name' column are in Arabic." Additionally, for datasets that only contain POIs from public spaces, we can point out the source of the dataset in the prompt, and amenities like ``toilets" will not be classified as ``home activity." Moreover, when dealing with real-world datasets where features often missing such as a POI labeled only with the amenity type ``visitor parking", it can be challenging to determine whether this POI belongs to the category ``visit friends" or ``pick up/drop off." For such cases, we can refine the prompt to guide the LLM in identifying the three most relevant categories for the POI, complete with probability scores for each. This tailored approach guides the model to provide meaningful classifications even with limited data, and these probabilities can be further utilized in other tasks, such as activity inference.

\subsection{Activity Inference}
Activity inference involves considering many factors, such as the time of the day, the category of the visited location, and the functionality of the location, requiring detailed information. However, the privacy implications of the use of location-based services~\cite{gruteser2003anonymous}, especially identifying specific POIs in the trajectory, are raising serious concerns. It could potentially disclose any specific user's daily activities, interests, and even personal habits. Thus, it is not appropriate to match the stay points in real trajectory data with specific POIs to conduct the inference. Therefore, an effective activity inference model must carefully balance the inclusion of detailed information with privacy safeguards. In the following parts, we elaborate a rule-based algorithm for mandatory activities inference and a probabilistic model for non-mandatory activities inference as shown in Fig. 3.

\textbf{\textit{Mandatory Activity Inference}}:
\begin{figure}[h]
    \centering
    \medskip
    \includegraphics[width=0.98\columnwidth]{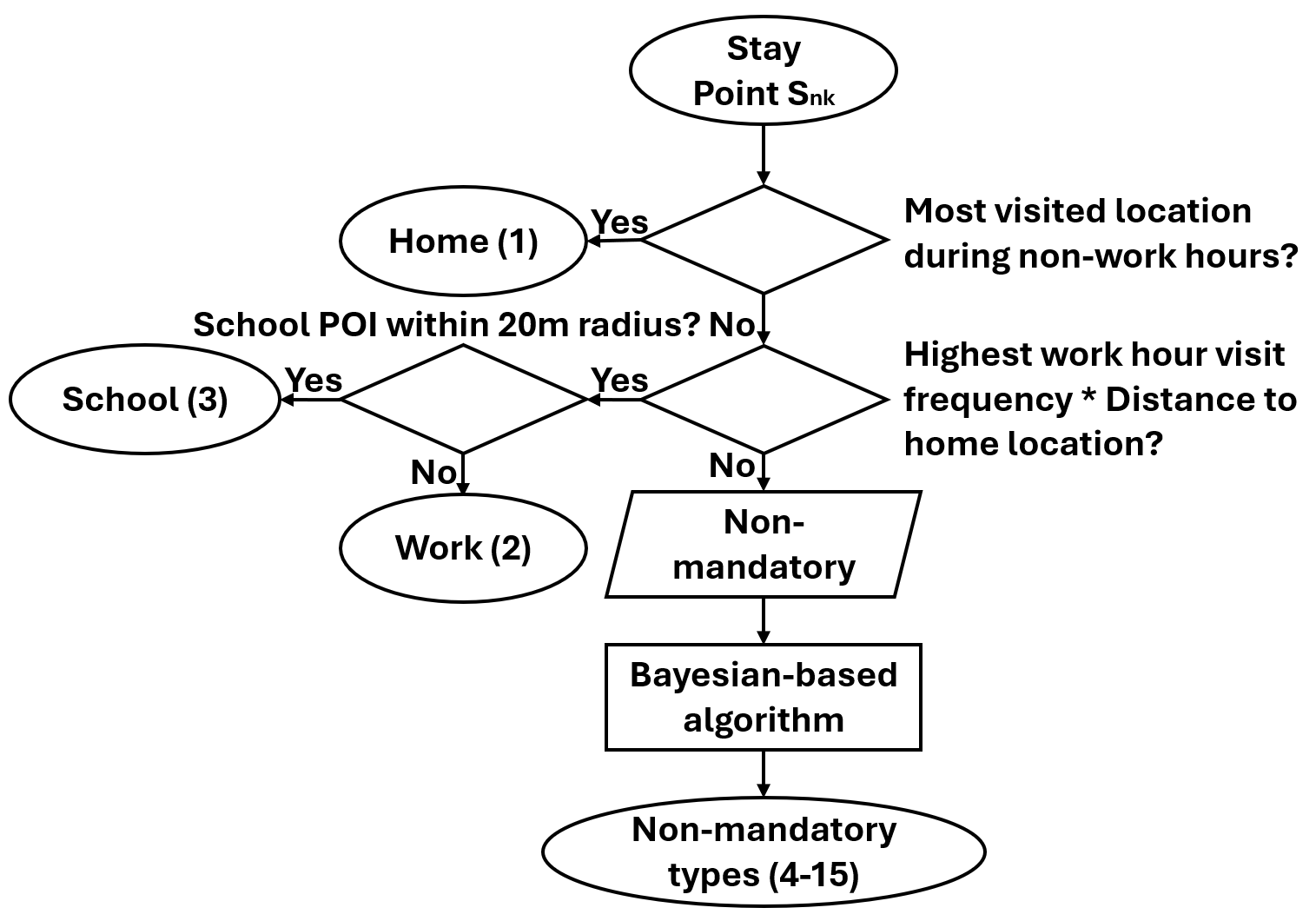}
    \caption{Flow chart of activity inference procedure.}
    \label{fig:prompt_example}
\end{figure}
Our mandatory activity inference algorithm identifies three primary activities: Home, Work, and School, using a rule-based approach \cite{alexander2015rule}. The Home activity is determined by the stay point with the highest visit frequency during off-hours (7 pm to 8 am). Work activity is inferred by analyzing stay points not marked as Home, focusing on those with the highest visit frequency and travel distance from Home during typical work hours on weekdays (8 am to 7 pm). For Schools, stay points closest to `education' POIs with the most travel from Home during weekdays' school hours are selected (8 am to 7 pm). This algorithm utilizes stay point data and proximity to relevant POIs to categorize mandatory activities effectively.


\textbf{\textit{Non-mandatory Activity Inference}}:

In contrast, non-mandatory activities, which vary more in periodicity and frequency, require a probabilistic approach for inference. We calculate the likelihood of activities at a stay point by considering the nearby \(K\) POIs, the noise-adjusted radius, and the start time $t_{S}$ of the stay.


The probability of an activity type $A_{m}$ can be formulated as the equation below, given the surrounding POIs $\mathcal{P} = \{p_1,..., p_k\}$ and the stay point start time $t_{S}$:

\begin{equation}
\begin{aligned}
P(A_m \mid \mathcal{P}, t_{S}) &= 
\sum_{K} P(A_m \mid p_k, t_{S}) P(p_k)
\end{aligned}
\end{equation}

Next, we use Bayes' theorem to refine the probability given $p_k$ and $t_{S}$:

\begin{equation}
P(A_m \mid p_k, t_{S}) = 
\frac{P(t_{S} \mid A_m) P(A_m \mid p_k)}
{P(t_{S} \mid p_k)}
\end{equation}

Substituting this back into our initial equation, we obtain the final expression to calculate the probability of a certain activity $A_m$:

\begin{equation}
\begin{aligned}
P(A_m \mid \mathcal{P}, t_{S}) &= 
\sum_{K} \frac{
P(t_{S} \mid A_m) 
}
{P(t_{S} \mid p_k)} P(A_m \mid p_k) P(p_k)
\end{aligned}
\end{equation}

Assuming the start time $t_{S}$ and a specific $p_k$ is independence, $P(t_{S} \mid p_k)$ becomes a constant value, and we can obtain:

\begin{equation}
\begin{aligned}
P(A_m \mid \mathcal{P}, t_{S}) &= 
\sum_{K} P(t_{S} \mid A_m) P(A_m \mid p_k) P(p_k)
\end{aligned}
\end{equation}

Each POI $p_k$ is assigned top three possible activity types $A_{1,2,3}$ and their probability $P(A_{1,2,3} \mid p_k)$ by LLM. $P(t_{S} \mid A_m)$ can be estimated from the 2017 National Household Travel Survey
(NHTS) California Add-on dataset \cite{NHTS2019}. The total conditional probability can be represented in a \(K \times 3\) matrix (as in Table \ref{table:poi_activity}), listing all possible probabilities for each POI and their corresponding activity types. The model then selects the highest probability among the possible POI and activity type combinations as the inferred activity result.

\section{Experiment}
\subsection{Dataset}
To evaluate our approach across various contexts and assess its effectiveness, we use open-source data from Los Angeles (LA) County and Egypt, sourced from OSM. This data enables us to rigorously test the performance of our POI classification algorithm.

The dataset is structured as a directed, connected network with POIs, each associated with metadata tags found on the network's edges or vertices. Given that OSM is a free, open geographic database maintained by volunteers through open collaboration, the data exhibits significant diversity in terms of quality, format, and variety. This includes a range of location-specific details such as buildings, amenities, and infrastructure elements like roads and railways. While it offers higher information entropy to the LLM, it also complicates predictions due to ambiguous or contradictory labels. For instance, a church in OSM could be labeled as a place of worship, a tourist attraction, and an office building simultaneously. While this might seem contradictory at first glance, such a POI is applicable in all three contexts, whether an agent is going for worship, tourism, or work.

Despite the crucial ``names" feature being present, 89\% of the content in other features such as ``amenity," ``building type," and ``land use" is missing in the Egypt dataset, and similarly, 91\% is missing in the LA dataset. Utilizing datasets like OSM enables us to evaluate the resilience of our methodology in challenging conditions due to the inconsistent labeling style and incomplete data in the real-world scenario.

\subsection{Evaluation Setting}
Our study covers 15 activity categories~\cite{NHTS2019} to capture a wide range of daily activities, ensuring the dataset reflects the diversity of human behavior, as detailed in Table \ref{table:activity_type_table}. 

\begin{table}[ht]
    \centering
    \caption{Activity category code and their corresponding descriptions}

    \begin{tabular}{|c|c|c|c|c|c|}
        \hline
        1 & Home & 2 & Work  & 3 & School \\ \hline
        4 & Caregiving & 5 & Buy goods & 6 & Buy services \\ \hline
        7 & Buy meals & 8 & General errands & 9 & Recreational \\ \hline
        10 & Exercise  & 11 & Visit friends & 12 & Health care \\ \hline
        13 & Religious & 14 & Something else & 15 & Drop off/Pick up\\ \hline
    \end{tabular}
    \label{table:activity_type_table}
\end{table}

\subsubsection{POI classification}

Evaluating the performance of our POI classification method presents challenges typical of an unsupervised learning setting, where no pre-existing ground truth is available. To address this, we randomly sample 500 observations from each POI dataset in both Egypt and LA County and create a manual ground truth. The accuracy of our method was then benchmarked against this human-labeled ground truth. To minimize labeling bias, five volunteers are enlisted to categorize the labels under uniform standards. Finally, we employ the language models `gpt-3.5-turbo (gpt3.5)' and `gpt-4 (gpt4)' from OpenAI, prompting them to identify the top three most relevant categories for each POI, along with the associated probabilities.




\subsubsection{Activity Inference}

After completing the POI classification using the `gpt-3.5' model, we obtain a dataset of 85,696 POIs, each annotated with possible activity categories, from LA County. This dataset serves as the input for the activity inference process, as illustrated in Fig. \ref{fig:overall_architecture}. 

To assess the efficacy of our proposed activity inference model, we curate a test trajectory dataset by extracting activities from the NHTS California Add-on dataset \cite{NHTS2019}, focusing on LA County. This dataset provides detailed daily travel trajectories, with each trajectory consisting of activities and a rough location at the zonal level. However, due to the lack of POIs associated with these activities, we utilize LLM for preliminary POI selection from the rough location zone, followed by manual curation to rectify any erroneous assignments. It is important to note the limitations of GPS-enabled smartphones, which typically provide accurate location data within a 4.9-meter radius under ideal conditions \cite{DOD2020GPS}. However, accuracy diminishes in urban environments near structures like buildings, bridges, and dense foliage. To evaluate the model's performance and robustness against real-world conditions, we introduce noise at three levels: 5 meters, 10 meters, and 20 meters. The noise, simulated using a Gaussian kernel, is added to the latitude and longitude coordinates of the POIs. Subsequent to this augmentation, our test dataset comprises 362 individuals and 2,007 activities, consisting of 1,724 mandatory activities and 283 non-mandatory activities. 

\subsection{Result}
\subsubsection{POI Classification}

\begin{figure*}[ht]
    \centering
    \medskip
    \includegraphics[width=1.0\linewidth]{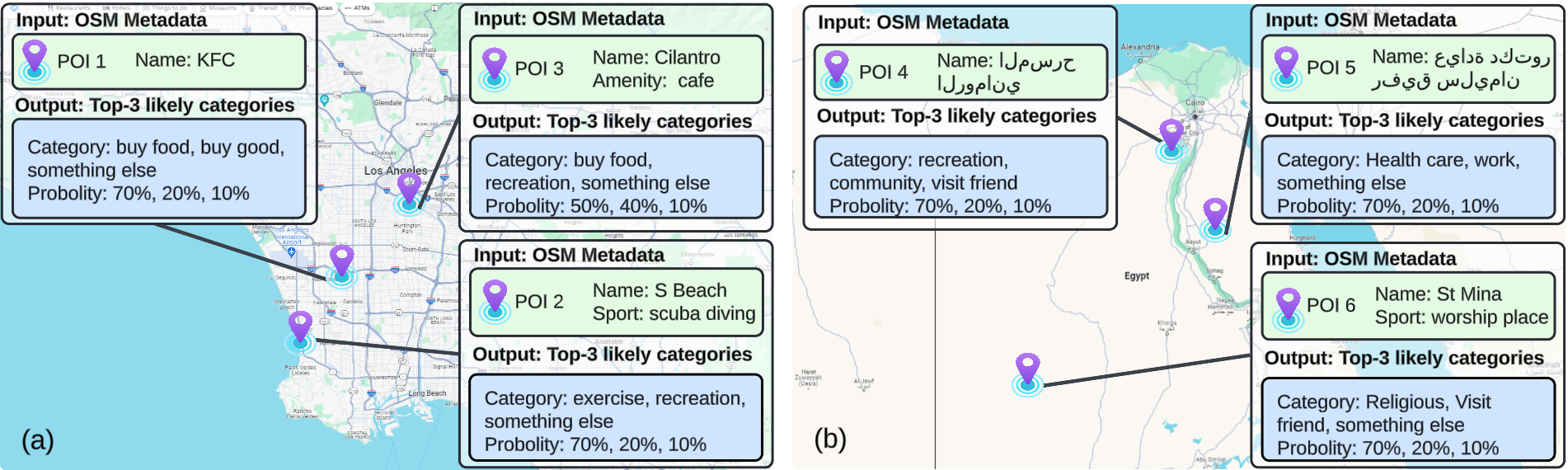}
    \caption{Example of POIs and classification results from (a) LA county  and (b) Egypt.}
    \label{fig:poi_example}
\end{figure*}

Considering there is a significant amount of incomplete features in the dataset. For observations that only have limited features available, such as a POI with only a name labeled like ``Volkswagen", it's hard to indicate whether the individual is there for car purchasing (category 5), auto maintenance (category 6), or is simply an employee at the location (category 2). As a result, in the calculation of overall accuracy and macro F-1 score, we consider the classification correct if one of the three predicted activities from the model matches the manually labeled activity category. Additionally, we calculate the hit rate for each position of the predicted activity code, recorded as ``Hit@n". The model returns these codes in decreasing order of probability, as specified in the prompt.

\begin{table}[ht]
    \centering
    \caption{Performance metrics for POI classification with different models.}
    \begin{tabular}{|c|c|c|c|c|}
        \hline
        \textbf{Metric}         & \textbf{Egypt(gpt3.5)} & \textbf{Egypt(gpt4)} & \textbf{LA(gpt3.5)} & \textbf{LA(gpt4)} \\
        \hline
        Accuracy          & 90.3\%                           & 93.4\%                         & 82.5\%                   & 91.4\%                 \\
        Hit @1          & 61.6\%                           & 74.7\%                         & 65.2\%                   & 75.7\%                 \\

        Hit @2          & 22.6\%                           & 9.7\%                         & 15.4\%                   & 8.7\%                 \\

        Hit @3          & 6.1\%                           & 9.0\%                         & 1.9\%                   & 7.0\%                 \\
        \hline
        F-1 Score               & 91.50\%                           & 96.10\%                         & 82.1\%                   & 92.9\%                 \\
        \hline

    \end{tabular}
    \label{table:poi_classification_matrix}
\end{table}

As shown in Table \ref{table:poi_classification_matrix}, our model achieves a 93.4\% accuracy for the Egypt dataset and 91.4\% accuracy for the LA county dataset, with F-1 scores of 96.10\% and 92.9\% respectively using the gpt4 model. This demonstrates its superiority and robustness compared with previous approaches. 

Compared to gpt3.5, the more advanced gpt4 model outperforms in both accuracy and F-1 score, especially for challenging datasets like LA country where more feature contents are missing. Besides, the analysis of the hit rate of the three predicted categories shows that while the ``Hit @1" rate dominates when using both models, the ``Hit @1" rate of gpt4 is approximately 10-13\% higher than that of gpt3.5. This improvement underscores the advanced model's enhanced comprehension abilities, leading to more robust performance in classification.



Some instances of the classification results are listed in Fig. \ref{fig:poi_example}. For POI 1, with only the name ``KFC", the model confidently categorized category 7 (Buy meals) as its first match with a probability of 0.7, and the semantically related ``buy" category 5 (Buy goods) as the second choice. For POI 2 with a clear description, the model predicted category 10 (Exercise) as the best match with a 0.7 probability, and category 9 (Recreational activities) as the next likely category with a 0.2 probability. In the case of POI 3, a cafe, the model assigned close probabilities between categories 7 (Buy meals) and 9 (Recreational), since a cafe usually includes both the dining and recreational social attributes. For POI 4 and POI 5 whose names were documented in Arabic(``Roman theater" and ``Dr. Rafiq Suleiman's clinic"), the model successfully understood the meaning and labeled them with corresponding categories. Besides, 14 (Something else) was chosen as the complementary third choice in all cases above as there are no other related categories. These impressing instances again demonstrate the strong comprehension ability of our approach, offering logically predicted probabilities that enhance subsequent activity inference tasks.



\subsubsection{Activity Inference}

In evaluating our activity inference model on the test dataset, we utilize accuracy @1, @2, and @3 under 5m, 10m, and 20m levels of noise as evaluation metrics. These metrics measure the model's performance in predicting activity types for stay points. Accuracy @1 represents the proportion of stay points where the correct activity type is the top prediction, while accuracy @2 and @3 indicate the percentages where the correct activity type falls within the top two and top three predictions, respectively.

Table \ref{table:poi_activity} provides examples of inference and demonstrate the probabilities of activity types and possible visiting POIs. In the example of clear inference, there are two possible POIs near the stay point in the trajectory data. By comprehensively considering the probability of categories and the time of the activity, the model is able to choose "Buy goods" as the possible activity category, which is consistent with the ground truth. Moreover, the probability of this correct category is significantly higher than in other cases. In contrast, in the example of ambiguous inference, the stay point is located in a commercial area mixed with many shopping and dining POIs. In such a scenario, the model finds "Buy goods" and "Buy meals" to be very close, making the situation confusing. The predicted probabilities of "Buy goods" and "Buy meals" are very close, leading to a prediction error. The model, even in ambiguous situations, can still make reasonable guesses, although in the case shown in the figure, the second choice is the most accurate inference. Overall, the model's inference can have strong distinguishability.

\begin{figure*}[t]
    \centering
    \medskip
    \includegraphics[width=1.0\linewidth]{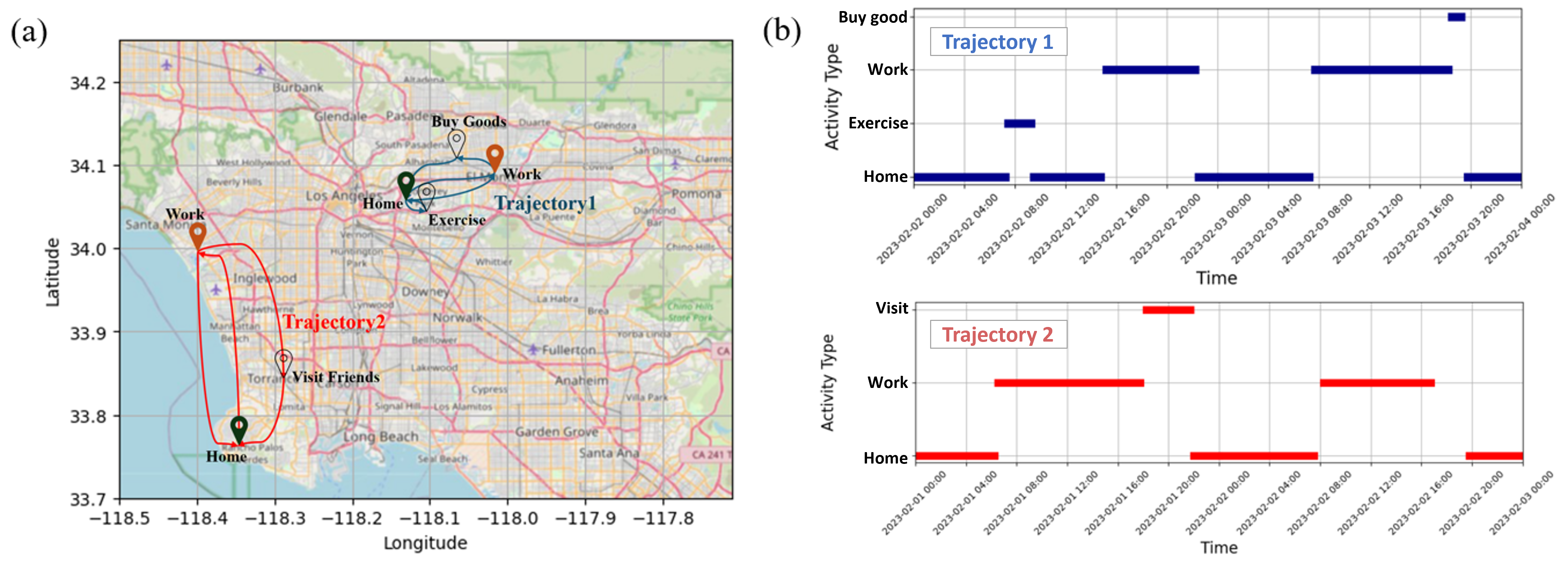}
    \caption{Examples of semantic trajectories post-data mining: (a) Activity-annotated stay points in LA. (b) Activity timelines.}
    \label{fig:Trajectory}
    \vspace{-3mm}
\end{figure*}

\begin{table}[h]
\centering
\caption{Conditional Probability of POI-Activity Combination for a Stay Point*}
\label{table:poi_activity}
\begin{adjustbox}{max width=\linewidth}
\begin{tabular}{|c|c|c|c|}
\hline
\multicolumn{4}{|c|}{An example of Clear Inference} \\ \hline
Activity for POI & $A_{1}$ & $A_{2}$ & $A_{3}$ \\ \hline
POI 1 (3.2m)** & \makecell{\textbf{0.424} \\ \textbf{(Buy goods)}} & \makecell{0.167 \\ (Buy services)} & \makecell{0.055 \\ (Something else)} \\ \hline
POI 2 (4.8m) & \makecell{0.214 \\ (Buy meals)} & \makecell{0.090 \\ (General errands)} & \makecell{0.047 \\ (Something else)} \\ \hline
\multicolumn{4}{|c|}{An example of Ambiguous Inference} \\ \hline
Activity for POI & $A_{1}$ & $A_{2}$ & $A_{3}$ \\ \hline
POI 1 (3.3m) & \makecell{0.170 \\ (Buy goods)} & \makecell{0.084 \\ (Buy services)} & \makecell{0.029 \\ (Something else)} \\ \hline
POI 2 (3.4m) & \makecell{0.169 \\ (Buy meals)} & \makecell{0.040 \\ (Buy goods)} & \makecell{0.012 \\ (Recreation)} \\ \hline
POI 3 (3.4m) & \makecell{0.182 \\ (Buy goods)} & \makecell{0.052 \\ (Buy services)} & \makecell{0.024 \\ (Recreation)} \\ \hline
POI 4 (4.2m) & \makecell{\textbf{0.208} \\ \textbf{(Buy meals)}} & \makecell{0.024 \\ (Recreation)} & \makecell{0 \\ (Buy goods)} \\ \hline
\end{tabular}
\end{adjustbox}
\begin{flushleft}
\small
* The activity inference algorithm identifies the most likely visited POI within a specified range, accounting for noise and multiple POIs, to determine the final activity type. \\
** Distance between the POI and the stay point.
\end{flushleft}
\end{table}

According to Table \ref{table:poi_classification_matrix_noise}, which outlines the performance metrics for activity inference under varying levels of noise, the algorithm maintains a commendable degree of stability even as the standard deviation (SD) of noise increases. Specifically, for non-mandatory activities, Acc @3 remains above 80\%, indicating a resilient performance against the perturbations caused by noise. With a 5-meter standard deviation in noise, the accuracy at Acc @3 is recorded at 88.4\%, which diminishes to 84.2\% when the noise level is increased to 20 meters. This reduction, while notable, is not drastic, supporting the assertion that the model upholds a stable predictive capability in the face of increasing noise levels. Particularly with respect to GPS trajectory datasets, the model demonstrates robust inference capabilities at a 5-meter noise accuracy. These findings not only validate the stability of the model but also highlight its potential for effective deployment in real-world scenarios where varying degrees of GPS accuracy are a common occurrence.

\begin{table}[ht]
    \centering
    \caption{Performance metrics for non-mandatory activity inference under different levels of noise}
    \begin{tabular}{|c|c|c|c|}
        \hline
        \textbf{SD of noise} & \textbf{Type Acc @1} & \textbf{Type Acc @2} & \textbf{Type Acc @3} \\
        \hline
        5m  & 75.0\% & 84.1\% & 88.4\% \\
        10m & 73.6\% & 81.8\% & 85.3\% \\
        20m & 71.9\% & 80.5\% & 84.2\% \\
        \hline
    \end{tabular}
    \label{table:poi_classification_matrix_noise}
    \vspace{-2mm}
\end{table}

As shown in Table \ref{table:activity_classification_matrix}, the results indicate that mandatory activities (categories 1-3) exhibit consistent performance due to the deterministic nature of their patterns. On the other hand, activities characterized by more complex and variable patterns present greater challenges for prediction. Despite this, categories 4, 5, 7, 8, 12, and 13 demonstrate commendable accuracy. Notably, categories such as 5 (Buy goods), 12 (Health care), and 13 (Religious activities) show robust performance, which can be attributed to the ample instances available within the test dataset.

Conversely, activities like 11 (Visit friends), 14 (Something else), and 15 (Drop off/Pick up) prove to be more elusive in the context of stay point detection. These activities are less associated with consistent POIs and often fall into the ambiguous situation as shown in Table~\ref{table:poi_activity}, leading to relatively poorer performance. This suggests that a rule-based algorithm is effective for activities with clear-cut patterns.

\begin{table}[ht]
    \centering
    \caption{Performance metrics for activity inference under 5m noise for different activity types}
    \small 
    \setlength{\tabcolsep}{2pt}
    \begin{tabular}{ccc|ccc}
        \toprule
        \textbf{Types} & \textbf{Acc@1} & \textbf{F1 score} & \textbf{Types} & \textbf{Acc@1} & \textbf{F1 score} \\
        \midrule
        1 & 98.3\% & 98.3\% & 9 & 63.1\% & 74.2\% \\
        2 & 87.2\% & 95.1\% & 10 & 64.7\% & 76.1\% \\
        3 & 90.0\% & 95.0\% & 11 & 52.2\% & 67.3\% \\
        4 & 100\% & 98.3\% & 12 & 100\% & 97.2\% \\
        5 & 78.5\% & 76.3\% & 13 & 100\% & 100\% \\
        6 & 66.7\% & 76.2\% & 14 & 46.2\% & 67.2\% \\
        7 & 77.4\% & 87.1\% & 15 & 50.1\% & 70.6\% \\
        8 & 100\% & 96.6\% & \textbf{Average} & 91.7\% & 92.3\% \\
        \bottomrule
    \end{tabular}
    \label{table:activity_classification_matrix}
    \vspace{-2mm}
\end{table}


Finally, the outcome of the proposed framework is the semantic trajectory, as illustrated in Fig. \ref{fig:Trajectory}, where each stay point in a trajectory is annotated with the most likely activity. For instance, as shown in Fig. \ref{fig:Trajectory} (a), individuals may engage in activities such as commuting between their home and work locations, exercising in the morning, or purchasing goods while en route from work to home. By augmenting stay point data with activity type information, we enhance our understanding of human travel mobility and behavior patterns. Through extensive trajectory mining, we progress from raw GPS data (aggregated to stay points) to refined activity chains, marking a significant advancement in the study of human travel behaviors.

\section{Conclusion and Future Work}
Our study has introduced a novel LLM-based framework for POI classification and activity inference, demonstrating notable improvements over traditional learning-based methods in handling incomplete and varied datasets like those from OpenStreetMap. Our approach, by leveraging the robust inferential capabilities of large language models, has shown exceptional performance with high accuracy and F-1 scores in real-world scenarios across diverse geographic locations. Looking ahead, we aim to enhance the accuracy and robustness of POI classification methods with more sophisticated prompt engineering or the development of a fine-tuned LLM designed specifically for the POI classification task.

\section{Acknowledgement}

We want to thank Xiangnan Zhang and Zhaoyi Ye for their contribution to dataset preprocessing.

\bibliographystyle{IEEEtran}
\bibliography{itsc_2024_llm_trajectory}

\begin{thebibliography}{10}
\providecommand{\url}[1]{#1}
\csname url@samestyle\endcsname
\providecommand{\newblock}{\relax}
\providecommand{\bibinfo}[2]{#2}
\providecommand{\BIBentrySTDinterwordspacing}{\spaceskip=0pt\relax}
\providecommand{\BIBentryALTinterwordstretchfactor}{4}
\providecommand{\BIBentryALTinterwordspacing}{\spaceskip=\fontdimen2\font plus
\BIBentryALTinterwordstretchfactor\fontdimen3\font minus \fontdimen4\font\relax}
\providecommand{\BIBforeignlanguage}[2]{{%
\expandafter\ifx\csname l@#1\endcsname\relax
\typeout{** WARNING: IEEEtran.bst: No hyphenation pattern has been}%
\typeout{** loaded for the language `#1'. Using the pattern for}%
\typeout{** the default language instead.}%
\else
\language=\csname l@#1\endcsname
\fi
#2}}
\providecommand{\BIBdecl}{\relax}
\BIBdecl

\bibitem{10480881}
A.~Petrillo and S.~Santini, ``Editorial special section on coordination, cooperation, and control of autonomous vehicles in smart connected road environments,'' \emph{IEEE Open Journal of Intelligent Transportation Systems}, vol.~5, pp. 202--204, 2024.

\bibitem{10356725}
L.~Kessler and K.~Bogenberger, ``Detection rate of congestion patterns comparing multiple traffic sensor technologies,'' \emph{IEEE Open Journal of Intelligent Transportation Systems}, vol.~5, pp. 29--40, 2024.

\bibitem{10131965}
R.~Chen, J.~Ning, Y.~Lei, Y.~Hui, and N.~Cheng, ``Mixed traffic flow state detection: A connected vehicles-assisted roadside radar and video data fusion scheme,'' \emph{IEEE Open Journal of Intelligent Transportation Systems}, vol.~4, pp. 360--371, 2023.

\bibitem{10354062}
J.~Ugan, M.~Abdel-Aty, and Z.~Islam, ``Using connected vehicle trajectory data to evaluate the effects of speeding,'' \emph{IEEE Open Journal of Intelligent Transportation Systems}, vol.~5, pp. 16--28, 2024.

\bibitem{pasquaretta2021analysis}
C.~Pasquaretta, T.~Dubois, T.~Gomez-Moracho, V.~P. Delepoulle, G.~Le~Loc’h, P.~Heeb, and M.~Lihoreau, ``Analysis of temporal patterns in animal movement networks,'' \emph{Methods in Ecology and Evolution}, vol.~12, no.~1, pp. 101--113, 2021.

\bibitem{10554659}
G.~Salierno, L.~Leonardi, and G.~Cabri, ``A big data architecture for digital twin creation of railway signals based on synthetic data,'' \emph{IEEE Open Journal of Intelligent Transportation Systems}, vol.~5, 2024.

\bibitem{shen2014review}
L.~Shen and P.~R. Stopher, ``Review of gps travel survey and gps data-processing methods,'' \emph{Transport reviews}, vol.~34, no.~3, 2014.

\bibitem{teusch2023systematic}
J.~Teusch, J.~N. Gremmel, C.~Koetsier, F.~T. Johora, M.~Sester, D.~M. Woisetschl{\"a}ger, and J.~P. M{\"u}ller, ``A systematic literature review on machine learning in shared mobility,'' \emph{IEEE Open Journal of Intelligent Transportation Systems}, vol.~4, pp. 870--899, 2023.

\bibitem{ruta2012semantic}
M.~Ruta, F.~Scioscia, S.~Ieva, G.~Loseto, and E.~Di~Sciascio, ``Semantic annotation of openstreetmap points of interest for mobile discovery and navigation,'' in \emph{2012 IEEE First International Conference on Mobile Services}.\hskip 1em plus 0.5em minus 0.4em\relax IEEE, 2012, pp. 33--39.

\bibitem{thonhofer2023infrastructure}
E.~Thonhofer, S.~Sigl, M.~Fischer, F.~Heuer, A.~Kuhn, J.~Erhart, M.~Harrer, and W.~Schildorfer, ``Infrastructure-based digital twins for cooperative, connected, automated driving and smart road services,'' \emph{IEEE Open Journal of Intelligent Transportation Systems}, 2023.

\bibitem{haklay2008openstreetmap}
M.~Haklay and P.~Weber, ``Openstreetmap: User-generated street maps,'' \emph{IEEE Pervasive computing}, vol.~7, no.~4, pp. 12--18, 2008.

\bibitem{choi2014poi}
S.~J. Choi, H.~J. Song, S.~B. Park, and S.~J. Lee, ``A poi categorization by composition of onomastic and contextual information,'' in \emph{2014 IEEE/WIC/ACM International Joint Conferences on Web Intelligence (WI) and Intelligent Agent Technologies (IAT)}, vol.~2.\hskip 1em plus 0.5em minus 0.4em\relax IEEE, 2014.

\bibitem{zhou2020poi}
C.~Zhou, H.~Yang, J.~Zhao, and X.~Zhang, ``Poi classification method based on feature extension and deep learning,'' \emph{Journal of Advanced Computational Intelligence and Intelligent Informatics}, vol.~24, 2020.

\bibitem{milias2021assessing}
V.~Milias and A.~Psyllidis, ``Assessing the influence of point-of-interest features on the classification of place categories,'' \emph{Computers, Environment and Urban Systems}, vol.~86, p. 101597, 2021.

\bibitem{zhao2023surveylargelanguagemodels}
\BIBentryALTinterwordspacing
W.~X. Zhao, K.~Zhou, J.~Li, T.~Tang, X.~Wang, Y.~Hou, Y.~Min, B.~Zhang, J.~Zhang, Z.~Dong, Y.~Du, C.~Yang, Y.~Chen, Z.~Chen, J.~Jiang, R.~Ren, Y.~Li, X.~Tang, Z.~Liu, P.~Liu, J.-Y. Nie, and J.-R. Wen, ``A survey of large language models,'' 2023. [Online]. Available: \url{https://arxiv.org/abs/2303.18223}
\BIBentrySTDinterwordspacing

\bibitem{bhattacharya2015matching}
T.~Bhattacharya, L.~Kulik, and J.~Bailey, ``Automatically recognizing places of interest from unreliable {GPS} data using spatio-temporal density estimation and line intersections,'' \emph{Pervasive and Mobile Computing}, vol.~19, pp. 86--107, 2015.

\bibitem{10382155}
D.~V. Cuong, V.~M. Ngo, P.~Cappellari, and M.~Roantree, ``Analyzing shared bike usage through graph-based spatio-temporal modeling,'' \emph{IEEE Open Journal of Intelligent Transportation Systems}, vol.~5, 2024.

\bibitem{alexander2015rule}
L.~Alexander, S.~Jiang, M.~Murga, and M.~C. González, ``Origin–destination trips by purpose and time of day inferred from mobile phone data,'' \emph{Transportation Research Part C}, vol.~58, 2015.

\bibitem{liu2023Graph}
X.~Liu, M.~Wu, B.~Peng, and Q.~Huang, ``Graph-based representation for identifying individual travel activities with spatiotemporal trajectories and poi data,'' \emph{Scientific Reports}, 2023, published on www.nature.com/scientificreports.

\bibitem{meng2017DBN}
C.~Meng, Y.~Cui, Q.~He, L.~Su, and J.~Gao, ``Travel purpose inference with gps trajectories, pois, and geo-tagged social media data,'' in \emph{2017 IEEE International Conference on Big Data (BIGDATA)}, 2017.

\bibitem{Jiang2016TimeGeo}
S.~Jiang, Y.~Yang, S.~Gupta \emph{et~al.}, ``The timegeo modeling framework for urban mobility without travel surveys,'' \emph{Proceedings of the National Academy of Sciences}, vol. 113, no.~37, pp. E5370--E5378, 2016.

\bibitem{gruteser2003anonymous}
M.~Gruteser and D.~Grunwald, ``Anonymous usage of location-based services through spatial and temporal cloaking,'' in \emph{Proceedings of the 1st International Conference on Mobile Systems, Applications, and Services (MobiSys)}, 2003, pp. 31--42.

\bibitem{NHTS2019}
\BIBentryALTinterwordspacing
{National Renewable Energy Laboratory}, ``{Transportation Secure Data Center},'' Accessed Jan. 15, 2019, 2019. [Online]. Available: \url{https://www.nrel.gov/tsdc}
\BIBentrySTDinterwordspacing

\bibitem{DOD2020GPS}
{U.S. Department of Defense}, ``{GPS} standard positioning service ({SPS}) performance standard,'' April 2020, 5th Edition.

\end{thebibliography}

\end{document}